\documentclass[lettersize,journal]{IEEEtran}
\usepackage{amsmath,amsfonts}
\usepackage{algorithmic}
\usepackage{algorithm}
\usepackage{array}
\usepackage{enumitem}
\usepackage[caption=false,font=normalsize,labelfont=sf,textfont=sf]{subfig}
\usepackage{textcomp}
\usepackage{stfloats}
\usepackage{xcolor}
\usepackage{hyperref}
\usepackage[table]{xcolor}
\usepackage{booktabs}
\usepackage{multirow}
\usepackage{url}
\usepackage{verbatim}
\usepackage{graphicx}
\usepackage{cite}
\newcommand{\reffig}[1]{{Fig.~\ref{#1}}}
\newcommand{\reftab}[1]{{Table~\ref{#1}}}
\newcommand{\refsec}[1]{{Section~\ref{sec:#1}}}

\begin{document}

\title{InterSing: Explicit Interaction Dynamics for 3D Duet Singing Animation and Beyond}

\author{ 	
        Yihan~Zhou,
        Zikai~Huang,
        Yuyang~Yu, 
 	      Xuemiao~Xu,
 	      Cheng~Xu,
 	      and~Shengfeng He,~\IEEEmembership{Senior Member,~IEEE}
\thanks{\textit{(Yihan Zhou and Zikai Huang contributed equally to this work.)}}
\thanks{Yihan Zhou, Zikai Huang, Yuyang Yu and Xuemiao Xu are with the School of Computer Science and Engineering, South China University of Technology, Guangdong, China.
     Xuemiao Xu is also with Guangdong Engineering Center for Large Model and GenAI Technology, and also with State Key Laboratory of Subtropical Building and Urban Science, Ministry of Education Key Laboratory of Big Data and Intelligent Robot. E-mail: 202521044545@mail.scut.edu.cn; 202210188523@mail.scut.edu.cn; 202410190026@mail.scut.edu.cn; xuemx@scut.edu.cn.}
\thanks{Cheng Xu and Shengfeng He are with the School of Computing and Information Systems, Singapore Management University. Email: cschengxu@gmail.com; shengfenghe@smu.edu.sg.}}
\markboth{}%
{Shell \MakeLowercase{\textit{et al.}}: A Sample Article Using IEEEtran.cls for IEEE Journals}


\maketitle

\begin{abstract}
We present InterSing, a framework for generating realistic 3D head animations for duet singing performances. Unlike solo singing, duet performance requires each singer to balance individual expressiveness with intermittent interaction at musically salient moments, such as phrase boundaries, synchronized rhythms, and call-and-response passages. Because these interactions are sparse and rhythm-dependent, existing audio-driven animation methods and conversational interaction models do not adequately capture their structure.
Our key insight is that duet coordination can be represented as a time-varying signal that reflects how strongly performers engage with one another throughout a song. Based on this observation, we introduce interaction logits, an interpretable latent representation that models the degree of cross-performer engagement at each time step. We learn these logits using weak supervision and use them to condition an interaction-aware diffusion model jointly driven by audio features and interaction dynamics. This formulation enables unified multi-mode generation, spanning independent motion, coordinated behavior, and smooth transitions between them.
Experiments show that InterSing generates realistic and expressive singing head animations with stronger coordination and musical alignment than existing methods, while preserving each performer's characteristic motion style. We further demonstrate that the same formulation generalizes to multi-singer performances and provides intuitive control over when and how performers engage.
Visual results and dataset are available on the \href{https://zhouyihann.github.io/InterSing/}{\textcolor{blue}{project page}}.
\end{abstract}


\section{Introduction}

  \begin{figure*}
    \includegraphics[width=\textwidth]{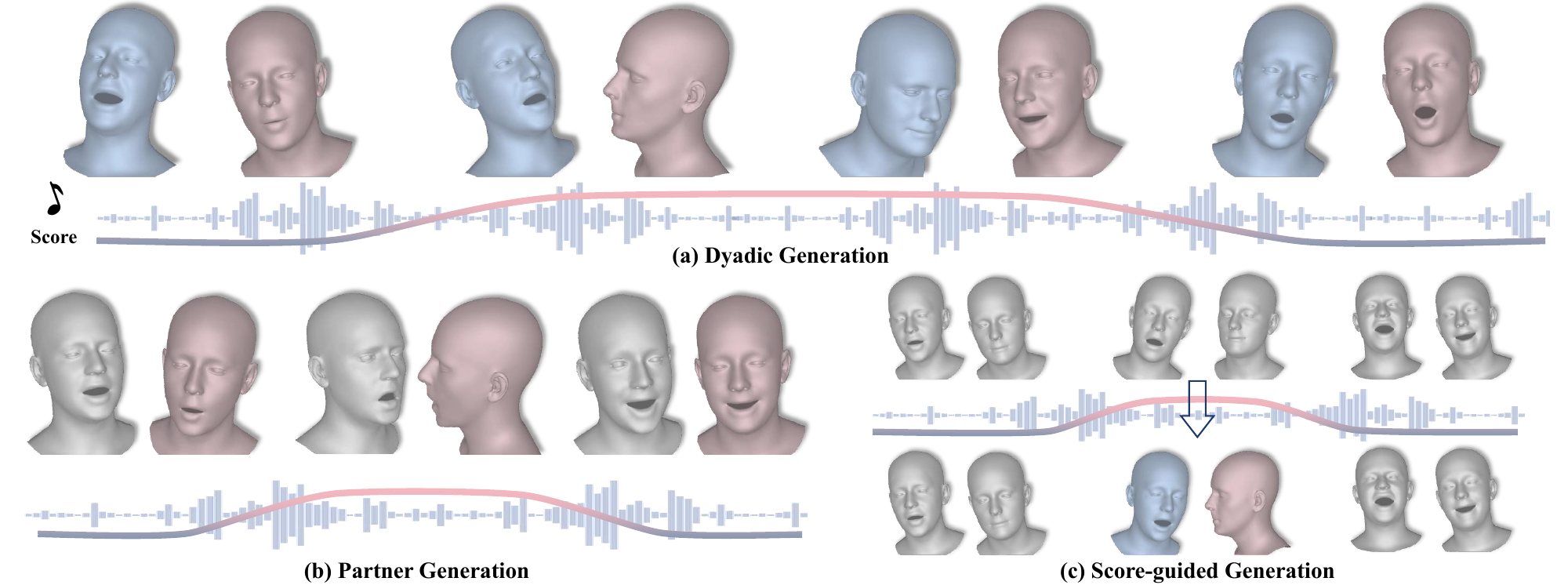}
    \caption{InterSing generates duet-singing animations from audio, with interaction scores either given or predicted from the same audio. The framework supports tasks including (a) dyadic generation, (b) partner generation, and (c) score-guided generation. In the partner generation and score-guided illustrations, the gray head denotes the given reference motion sequence.}
    \label{fig:teaser}
  \end{figure*}

\IEEEPARstart{R}{ealistic} digital human animation plays a central role in modern virtual production, enabling applications such as virtual concerts, digital performers, and immersive entertainment~\cite{yu20193d, huang2025think2sing, xie2025let, liu2024musicface, wu2025singinghead}. Among these scenarios, singing performance presents a particularly challenging setting for motion synthesis. Unlike conventional speech-driven animation, singing requires the coordinated generation of facial expressions, head movements, and rhythmically aligned motion patterns governed by musical structure~\cite{quinto2013emotional, livingstone2013acoustic, eyben2015emotion, livingstone2015common, pan2022vocal, el2011survey}. These expressive behaviors extend beyond phonetic articulation and demand temporally coherent dynamics that reflect both musical timing and performer expressiveness.

Recent audio-driven 3D head animation methods have achieved promising results for single performers~\cite{chu2025artalk, nocentini2024scantalk, fan2024unitalker, fan2022faceformer, wu2025singinghead}. However, singing often occurs as a shared performance. In duet singing, each performer must preserve individual expressiveness while intermittently coordinating with a partner. Such coordination appears through subtle but perceptually important cues, including head orientation, facial affect, rhythmic emphasis, and synchronized motion at phrase boundaries, choruses, or call-and-response passages. These moments create the sense of mutual awareness and stage presence that distinguishes a coherent duet from two independently animated singers.
Duet interaction also differs fundamentally from conversational interaction. In dialogue, coordination is often continuous, with participants repeatedly exchanging gaze, nods, and backchannel feedback. In singing, by contrast, interaction is sparse and structured by music. Performers may remain visually and expressively independent for long intervals, then briefly align at musically salient moments. Moreover, these interactions are not merely reactive to another person's motion, but are jointly driven by rhythm, phrasing, and performance structure. As a result, interaction models designed for conversational animation do not directly capture the intermittent and rhythm-aware nature of duet singing.

Existing dyadic motion generation methods~\cite{peng2026dyaditmultimodaldiffusiontransformer, chu2025unils, tran2024dim, peng2025dualtalk, ng2022learning, xie2025let} typically rely on joint latent spaces or shared generation processes. While such designs can synthesize two-person motion, they rarely expose the underlying interaction state between performers. Without an explicit representation of when and how strongly performers engage, generated results may either behave as two independent solo animations or exhibit overly coupled motion. This also limits controllability in production workflows, where artists may wish to adjust the timing, frequency, or intensity of interaction without resynthesizing individual performance styles.

To address these challenges, we formulate \textit{duet-singing head animation} as an interaction-conditioned motion generation problem. Our key idea is to separate performer motion from interaction dynamics by introducing \textit{interaction logits}, a time-varying scalar representation that describes the degree of coordination between singers. The interaction logits act as an interpretable and editable control signal, enabling the model to generate independent motion, coordinated duet behavior, and smooth transitions between them within a unified framework, as shown in ~\reffig{fig:teaser}.

Learning such a representation is non-trivial because interaction in duet singing is difficult to annotate and inherently subjective. We therefore propose a weakly supervised interaction learning framework based on contrastive representation learning~\cite{chen2020simple, oord2018representation, khosla2020supervised, zheng2021weakly}. From paired FLAME motion sequences, we derive coarse geometric anchors using relative head orientation, motivated by the observation that performers often turn toward each other during moments of engagement. These anchors provide useful but incomplete supervision: some interactions occur without clear head orientation changes, while similar poses may arise without genuine coordination. We thus treat them not as ground truth labels, but as sparse cues that guide representation learning.
However, this geometric heuristic provides only a partial and noisy observation of true interaction dynamics. Many interaction events may occur without explicit head orientation changes, while similar orientations may arise without meaningful coordination. Consequently, the resulting signal inevitably contains both false negatives and false positives. Rather than treating these labels as ground-truth annotations, we interpret them as sparse anchors that provide an initial inductive bias for interaction representation learning. To further prevent the model from relying on trivial geometric shortcuts, we introduce a neck-masking strategy that occludes neck-related pose parameters during interaction encoding. This design prevents the encoder from directly inferring interaction from head orientation parameters and encourages it to capture richer coordination cues from facial dynamics and rhythmic motion patterns. Through contrastive learning with multiple negative sampling strategies, the model aggregates consistent temporal patterns across sequences and refines these sparse anchors into a temporally coherent interaction representation.

Building on the learned interaction representation, we propose an interaction-aware diffusion transformer for duet motion generation. The model conditions jointly on audio features and interaction logits to synthesize coordinated dyadic head motions. We further adopt diffusion forcing~\cite{chen2024diffusion} to train the model under multiple noise conditions, allowing a single framework to support several practical modes: full duet synthesis, interaction-aware refinement of existing solo performances, and partner motion prediction.
Extensive experiments show that explicit interaction modeling improves the realism, musical alignment, and perceived coordination of duet-singing animation. Our method generates expressive dual-performer motions while preserving each singer's characteristic motion style. We also demonstrate that the learned interaction logits provide intuitive control over interaction timing and intensity, and that the formulation naturally extends to multi-singer performances.
In summary, our contributions are threefold:
\begin{itemize}[leftmargin=*]
    \item We formulate duet-singing head animation as an interaction-conditioned motion generation problem, emphasizing the intermittent and rhythm-driven coordination patterns of singing performance.

    \item We introduce a weakly supervised interaction representation learning framework that uses geometric prior anchors, neck-occluded encoding, and contrastive learning to infer robust interaction logits from paired motion sequences.

    \item We propose an interaction-aware diffusion generation framework with diffusion forcing, enabling unified and controllable multi-mode synthesis for duet-singing head animation.
\end{itemize}

\section{Related Work}
\subsection{Audio-Driven 3D Head Animation}

Audio-driven 3D head animation has long been an active research area in computer graphics and animation. Early approaches \cite{edwards2016jali, taylor2012dynamic, xu2013practical} mainly relied on rule-based systems that segmented speech into phonemes and mapped them to predefined lip-synchronized animations. Because these methods depended heavily on handcrafted rules, they were limited in capturing subtle variations in human expression. More recently, deep learning has significantly advanced audio-driven 3D facial animation. Existing speech-driven methods \cite{fan2022faceformer,peng2023selftalk,xing2023codetalker,nocentini2024scantalk, chu2025artalk, fan2024unitalker,stan2023facediffuser,thambiraja2023imitator, peng2023emotalk, danvevcek2023emotional, song2024expressive} generate realistic facial motions using mesh or vertex representations and show strong cross-identity generalization for retargeting. However, these methods are mainly designed for general speech scenarios and remain less effective for singing, where capturing distinctive prosodic patterns and fine-grained emotional expressions is particularly important.

To address this limitation, several singing-driven facial animation methods have recently been introduced. SingingHead \cite{wu2025singinghead} and MusicFace \cite{liu2024musicface} model pitch variation and emotional transition from acoustic and ASR features within audio-driven frameworks, producing more natural singing animations. Rather than relying only on low-level audio features, Think2Sing \cite{huang2025think2sing} uses LLMs to infer fine-grained facial motion descriptions from both lyrics and audio signals, enabling expressive animations that better reflect lyrical semantics. 
Although these methods achieve promising results for single-character animation, they are not designed for multi-character interaction. As a result, they often struggle to maintain emotional synchronization across characters and to generate naturally coordinated head motions.

\subsection{Audio-Driven 3D Interactive Head Animation}

In recent years, audio-driven 3D head animation has evolved from single-person generation toward multi-person interactive scenarios, with speak--listen facial animation becoming an important research direction. In conversational settings, early studies~\cite{wang2025diffusion,cai2025flooddiffusion,liu2023mfr,liu2024customlistener,ng2023can,ng2024audio,zhou2022responsive,tran2024dim,song2023emotional} mainly focused on modeling nonverbal listener behaviors. For example, Learning2Listen~\cite{ng2022learning} uses a unidirectional conditional generation framework to synthesize listener facial responses solely from the speaker's audio and facial motions. Later methods~\cite{liu2024customlistener,tran2024dim,song2023emotional} incorporate conversational context into expression prediction, but they still rely on decoupled short-term response modeling. As a result, these methods remain limited to isolated single-turn interactions and struggle to capture the fluid bidirectional interactions required in long conversations.

DualTalk~\cite{peng2025dualtalk} partially addresses this limitation by jointly generating speaking and listening facial motions. However, it requires precomputed interlocutor motion sequences, which limits end-to-end learning and reduces applicability in real-time scenarios. UniLS~\cite{chu2025unils} removes explicit facial motion inputs and directly models interactive facial animation from dual-channel speech audio, based on the assumption that listener behavior is governed by an audio-modulated internal motion prior. Although these methods perform well in conversational settings, their interaction paradigm differs fundamentally from duet singing. Conversations usually involve relatively stable interaction patterns with fixed speaker-listener roles, whereas duet singing involves intermittent, highly collaborative interactions with frequent role transitions and tightly coordinated movements. This structural difference makes existing speech-listening frameworks difficult to transfer directly to singing scenarios.

Recently, PaChorus~\cite{xie2025let} introduced the first method for duet-singing head animation generation based on joint motion synthesis. However, the method models inter-singer interactions only implicitly, which often leads to limited motion coordination and weak temporal alignment between performers. In contrast, our work formulates duet-singing head animation generation as an interaction-conditioned motion synthesis problem with explicit modeling of cross-performer relationships. The proposed interaction representation is interpretable and controllable, enabling fine-grained modulation of inter-singer dynamics while preserving each performer's individual singing characteristics.

\begin{figure*}[t]
  \centering
  \includegraphics[width=\linewidth]{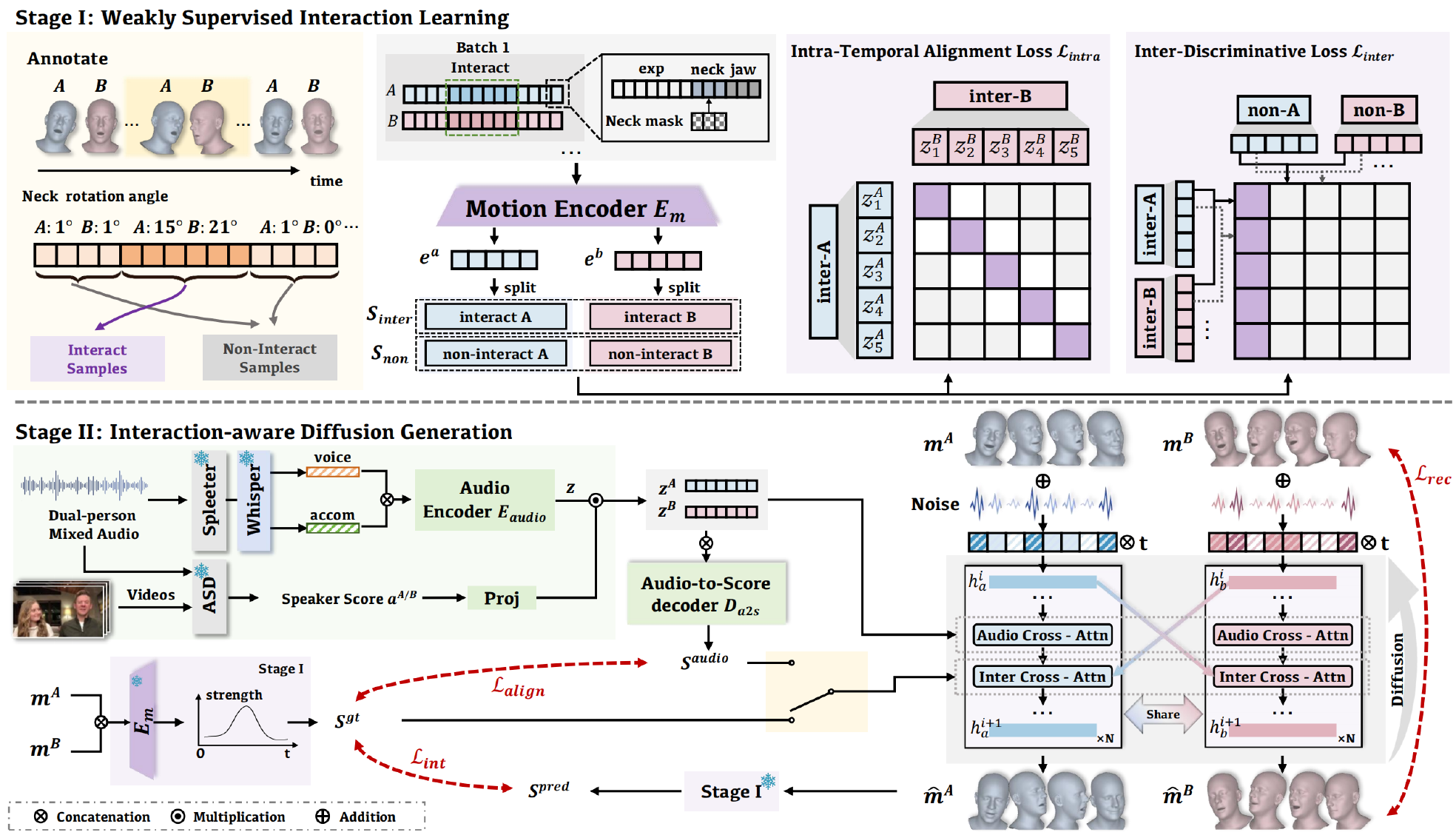}
    \caption{
    \textbf{Overview of InterSing.} 
    Stage~I partitions motion encoder $E_m$ embeddings into interaction $\mathcal{S}_{\text{inter}}$ and non-interaction $\mathcal{S}_{\text{non}}$ sets via weakly-supervised geometric priors. Intra-Temporal Alignment Loss $\mathcal{L}_{intra}$ maximizes \textcolor[RGB]{112,48,160}{temporally aligned positive pairs} against \textcolor{gray}{temporally distant negatives} to enforce temporal coherence, while Inter-Discriminative Loss $\mathcal{L}_{inter}$ maximizes \textcolor[RGB]{112,48,160}{interaction positive pairs} against \textcolor{gray}{non-interaction negatives} to sharpen the interaction decision boundary. Stage~II injects the learned interaction scores as explicit conditioning signals into a Diffusion Forcing framework with cascaded cross-attention, jointly denoising both singers' motions under variable noise horizons that enable flexible generation modes.
    }
  \label{overview}
\end{figure*}

\section{Method}
An overview of our method is presented in~\reffig{overview}.
We employ the widely used 3D Morphable Model FLAME~\cite{li2017learning} to obtain a parameterized 3D representation of the head, where facial expressions, neck pose, and jaw pose are respectively represented by the parameters $\boldsymbol{\psi} \in \mathbb{R}^{|\psi|}$, $\boldsymbol{\theta}^{\text{neck}} \in \mathbb{R}^{3}$, and $\boldsymbol{\theta}^{\text{jaw}} \in \mathbb{R}^{3}$. 
Based on this representation, we propose a two-stage framework InterSing to capture the intermittent interaction dynamics in singing performances.
The framework consists of weakly supervised interaction learning followed by interaction-aware diffusion-based generation.
\refsec{stage1} details our weakly supervised interaction learning method. \refsec{stage2} introduces the interaction-aware diffusion generation framework. Finally, \refsec{inference} presents various inference strategies for generating duet-singing head animations based on the learned interaction cues and the diffusion generation framework.

\subsection{Stage I: Weakly Supervised Interaction Learning} \label{sec:stage1}
Modeling performer interaction in duet singing is challenging due to the scarcity of reliable frame-level annotations.
We observe that in duet singing, moments of high-intensity interaction often correspond to performers orienting their heads toward each other, while during less coordinated passages, head orientations are more independent. Motivated by this observation, we derive frame-level pseudo-labels from the relative neck orientations of the singers. 
Let $\phi$ denote the mutual neck rotation angle and $\eta$ the predefined threshold. Frames satisfying $\phi > \eta$ are regarded as interaction samples, while the remaining frames are assigned to non-interaction samples. These sparse and noisy anchors provide a guiding signal for learning interaction representations without imposing strict ground-truth constraints.

To capture temporal coordination patterns beyond these coarse cues, we employ a Transformer-based motion encoder $E_m$ that maps duet motion sequences $\mathbf{m}^A, \mathbf{m}^B \in \mathbb{R}^{L \times D_m}$ into a joint latent space, where $L$ is the number of frames and $D_m$ denotes the motion dimension. 
However, directly learning from the original motion representation may introduce a potential shortcut, as the pseudo-labels are partially derived from mutual neck orientations. 
The encoder could therefore exploit explicit neck rotations to distinguish interaction and non-interaction states, rather than learning the underlying coordination patterns between performers.
To mitigate trivial geometric shortcuts, we introduce a neck-masking strategy during training, where 40\% of the neck pose features are randomly masked for each motion sequence.
Instead of relying solely on explicit head orientation cues, the encoder is encouraged to infer interaction states from complementary signals, including subtle facial expression changes, synchronized head movements, and temporal dependencies between performers.
As a result, the learned embeddings capture higher-level interaction semantics beyond direct spatial alignment, providing a more robust representation for subsequent interaction-aware generation.
The encoded sequences pass through dual projection heads to produce individual embeddings $e^A, e^B \in \mathbb{R}^{L \times D_e}$, where $D_e$ is the embedding dimension. Guided by pseudo-labels derived from interaction and non-interaction samples, $e^A, e^B$ are partitioned into interaction and non-interaction sets, denoted as $\mathcal{S}_{inter}$ and $\mathcal{S}_{non}$:
\begin{equation}
\begin{aligned}
\mathcal{S}_{inter} &= \{ (e_i^A, e_i^B) \mid \phi_i > \eta \},\\
\mathcal{S}_{non}  &= \{ (e_i^A, e_i^B) \mid \phi_i \leq \eta \}.
\end{aligned}
\end{equation}
where $e_i^A, e_i^B \in \mathbb{R}^{D_e}$ are the embeddings of performers $A$ and $B$ at frame $i$.
The embeddings are optimized with two complementary contrastive objectives. The intra-temporal alignment loss $\mathcal{L}_{intra}$ encourages consistency of interaction embeddings within the same temporal frame, while the inter-discriminative loss $\mathcal{L}_{inter}$ enforces separation between interaction and non-interaction pairs across frames. 

\textbf{Intra-Temporal Alignment Loss $\mathcal{L}_{intra}$.} 
To capture temporal coherence of interaction dynamics, we introduce a weakly supervised temporal interaction loss. Within the interaction set $\mathcal{S}_{inter}$, embedding pairs $(e_i^A, e_i^B)$ from the same frame are treated as positives, while pairs separated by more than $d$ frames ($|i-k|>d$) serve as negatives. Temporally close pairs are excluded to avoid separating highly similar frames. This formulation encourages temporally aligned interaction embeddings to be close while pushing misaligned ones apart, yielding a discriminative and temporally coherent representation of dyadic coordination:
\begin{equation}
\begin{aligned}
p_{intra,i}
&=
\frac{\exp(e_i^A \cdot e_i^B / \tau)}
{
\sum_{\substack{k \in \mathcal{S}_{inter}\\ |i-k|>d}}
\exp(e_i^A \cdot e_k^B / \tau)
+
\exp(e_i^A \cdot e_i^B / \tau)
}
\end{aligned}
\end{equation}

\begin{equation}
\mathcal{L}_{intra} = -\frac{1}{N_{inter}} \sum_{i=1}^{N_{inter}} \log p_{intra, i} ,
\end{equation}
where $N_{inter}$ denotes the number of interaction samples in $\mathcal{S}_{inter}$, $k$ indexes this set, $d$ is the temporal distance threshold, and $\tau$ is the temperature hyperparameter. By leveraging the weak pseudo-labels derived from geometric cues, this loss refines sparse interaction anchors into a robust temporal embedding.

\textbf{Inter-Discriminative Loss $\mathcal{L}_{inter}$.} 
To further enforce separation between interaction and non-interaction states, we define an interaction discrimination loss. Simultaneous embedding pairs from $\mathcal{S}_{inter}$ are treated as positives, while those from the non-interaction set $\mathcal{S}_{non}$ are treated as negatives. This loss encourages the model to pull together embeddings of interacting frames and push apart non-interacting ones, ensuring that the latent space captures meaningful dyadic coordination beyond trivial geometric correlations:
\begin{equation}
p_{inter,i} = \frac{\exp(e_i^A \cdot e_i^B / \tau)}
{\sum_{k \in \mathcal{S}_{non}} \exp(e_k^A \cdot e_k^B / \tau)
+ \exp(e_i^A \cdot e_i^B / \tau)}
\end{equation}

\begin{equation}
\mathcal{L}_{inter} = -\frac{1}{N_{inter}} \sum_{i=1}^{N_{inter}} \log p_{inter,i} .
\end{equation}

The total loss is $\mathcal{L} = \mathcal{L}_{intra} + \mathcal{L}_{inter}$. 
By combining these objectives, the model transforms sparse geometric anchors into a robust, temporally coherent, and interpretable representation that captures semantically meaningful dyadic interaction patterns, forming a strong foundation for subsequent interaction-aware singing animation.

\subsection{Stage II: Interaction-aware Diffusion Generation} \label{sec:stage2}
Building upon the interaction encoder ${E}_{m}$ pretrained in the first stage, we aim to synthesize temporally coherent, coordinated duet-singing head animations. A key challenge lies in the inherently asymmetric and intermittent nature of duet interaction, where the degree of coordination varies over time, conditioned jointly on audio rhythm and interaction intensity.
This motivates \textit{InterSing}, a Transformer-based diffusion model that jointly generates dyadic singing head motions by incorporating audio and interaction-aware control signals directly into the denoising process, enabling temporally coherent and controllable coordination between performers.
Raw audio is first separated into vocal and accompaniment tracks using Spleeter~\cite{spleeter2020}, each track is encoded by a frozen Whisper encoder~\cite{radford2023robust}, and the resulting embeddings are concatenated and processed by an Audio Transformer $E_{audio}$ to produce the input latent representation $\mathbf{z}$.
Frame-wise speaker active signals $\mathbf{a}^{p} \in [0,1]^L$ are produced by an Active Speaker Detection module (LR-ASD)~\cite{liao2025lr} and fused with the audio features $\mathbf{z}$ to obtain speaker-aware representations $\mathbf{z}^p$ for each singer $p \in \{A, B\}$.
Unlike conventional audio-driven animation that models each performer independently, we introduce interaction intensity as an explicit latent variable to regulate the coordination strength between two motion streams.
Dyadic coordination is explicitly modeled using interaction signals from the pretrained encoder ${E}_{m}$. Frame-level embedding similarities between paired motions $(\mathbf{m}^A, \mathbf{m}^B)$ define pseudo ground-truth scores $s^{gt}$, providing a continuous signal to control cross-person coordination.
Specifically, the cosine similarity between paired interaction embeddings is used as the interaction score:
\begin{equation}
s_i^{gt} =
\frac{1}{2}
\left(
\frac{e_i^A \cdot e_i^B}
{\|e_i^A\|_2\|e_i^B\|_2}
+1
\right).
\end{equation}
where $s_i^{gt}\in[0,1]$ represents the continuous interaction intensity at frame $i$.
During training, the interaction signal is randomly sampled either from $s^{gt}$ or predicted by an audio-to-score decoder $D_{a2s}$.
Since $s^{gt}$ is extracted from motion embeddings, it provides reliable interaction supervision during training but is unavailable at inference time when only audio input is provided. 
To bridge this gap, we train an auxiliary audio-to-score decoder to estimate interaction dynamics from audio cues.
Although interaction is primarily expressed through visual coordination, singing audio still contains complementary information, such as rhythmic structure and phrasing patterns, that correlates with performer coordination.
$D_{a2s}$ takes the concatenated speaker-aware audio features from both singers as input and predicts interaction scores $s^{audio}$ via a shallow Transformer.
Inspired by previous works~\cite{maluleke2025diffusion, petrov2025echo, cai2025flooddiffusion, wu2025uniphys}, we instantiate the diffusion process in the Diffusion Forcing setting~\cite{chen2024diffusion}, which extends DDPM~\cite{ho2020denoising} by introducing token-wise stochasticity to enable flexible temporal modeling.
For each singer $p$ and frame $i$, we define the noisy motion token at diffusion step $t_i^p$ as:
\begin{equation}
m_i^p(t_i^p) = \sqrt{\alpha(t_i^p)} \, m_i^p + \sqrt{1-\alpha(t_i^p)} \, \boldsymbol{\epsilon}_i^p, 
\quad \boldsymbol{\epsilon}_i^p \sim \mathcal{N}(\mathbf{0}, \mathbf{I}),
\end{equation}
where $m_i^p$ is the ground-truth motion for frame $i$ of singer $p$, $t_i^p$ is the token-specific diffusion timestep, and $\boldsymbol{\epsilon}_i^p$ is independently sampled Gaussian noise. Each $t_i^p$ is drawn independently across frames and singers, allowing heterogeneous corruption across both temporal and motion dimensions. This token-wise noise formulation provides a unified stochastic space that can represent both tightly synchronized and loosely coupled motion segments.

The generation is formulated as a conditional diffusion process that jointly denoises $\mathbf{m}^{A}(\mathbf{t}^A)$ and $\mathbf{m}^{B}(\mathbf{t}^B)$ using a Transformer with cascading attention, conditioned on token-wise diffusion timesteps $\mathbf{t}^{A}$ and $\mathbf{t}^{B}$, speaker-aware audio features $\mathbf{z}_p$, and interaction signals.
To capture mutual dependencies between the two singers, the network further exploits cross-stream hidden states from the counterpart branch via interaction cross-attention~\cite{liang2024intergen}.
The noisy motion inputs are projected into a shared latent space and encoded into initial hidden states $h_{A}^{(0)}$ and $h_{B}^{(0)}$, which are iteratively refined through $N$ stacked attention blocks and finally projected to reconstruct the denoised motions $\hat{\mathbf{m}}^A$ and $\hat{\mathbf{m}}^B$. 
Each block integrates self-attention to preserve temporal coherence, audio cross-attention to inject speech-driven dynamics via $\mathbf{z}_p$, and interaction cross-attention~\cite{liang2024intergen} to model dyadic coupling. To enable controllable coordination, FiLM-based modulation~\cite{perez2018film} applies interaction signals to the cross-attention layers and injects the diffusion timestep $t$ for noise-aware generation.

The model is trained to enforce both motion fidelity and dyadic coordination. Motion fidelity is encouraged through three kinetic losses:
\begin{equation}
\begin{aligned}
\mathcal{L}_{rec} &= \|\hat{\mathbf{m}} - \mathbf{m}\|_2^2, \\
\mathcal{L}_{vel} &= \|\hat{\mathbf{m}}' - \mathbf{m}'\|_2^2, \\
\mathcal{L}_{acc} &= \|\hat{\mathbf{m}}'' - \mathbf{m}''\|_2^2.
\end{aligned}
\end{equation}
These terms jointly enforce accurate and temporally smooth motion.
Dyadic coordination is enforced through two interaction-level losses:
\begin{equation}
\begin{aligned}
\mathcal{L}_{align} &= \| s^{audio} - s^{gt} \|_2^2, \\
\mathcal{L}_{int} &= \| \hat{s} - s^{gt} \|_2^2,
\end{aligned}
\end{equation}
where $\hat{s}$ is the interaction signal extracted from the generated motion $\hat{\mathbf{m}}$ via the interaction encoder $E_m$.
The alignment loss $\mathcal{L}_{align}$ enforces consistency between the audio-conditioned prediction and the pseudo ground-truth, enabling the model to infer interaction dynamics from audio alone during inference. 
The interaction consistency loss $\mathcal{L}_{int}$ further constrains the generated motions by matching the interaction intensity extracted from synthesized motions with the desired interaction score.
Together, these objectives prevent the model from generating individually plausible but mutually independent motions, encouraging the two streams to exhibit coordinated behaviors consistent with the desired interaction intensity.
The overall objective is a weighted sum of all terms, expressed as $\mathcal{L}_{total} = \lambda_1 \mathcal{L}_{rec} + \lambda_2 \mathcal{L}_{vel} + \lambda_3 \mathcal{L}_{acc} + \lambda_4 \mathcal{L}_{align} + \lambda_5 \mathcal{L}_{int}$, where $\lambda_1, \dots, \lambda_5$ balance the contribution of each component.

\subsection{Inference} \label{sec:inference}
\begin{figure}[t]
  \centering
  \includegraphics[width=\linewidth]{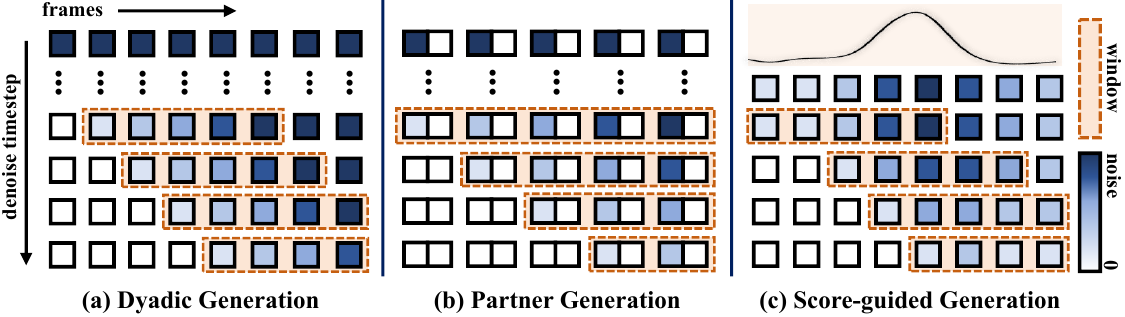}
  \caption{
  \textbf{Flexible Inference via Diffusion-Forcing.} Our framework enables (a) dyadic generation, (b) partner generation, and (c) score-guided generation through different noise scheduling}
  \label{inference}
\end{figure}
InterSing provides a unified framework for controllable duet-singing head animation, combining an interaction-aware generation paradigm with a diffusion-forcing training strategy. Instead of requiring separate models for different applications, a single architecture supports multiple inference modes by simply adjusting the noise initialization and conditioning strategy. This unified formulation enables flexible deployment while preserving temporally coherent and interaction-consistent motion generation. As illustrated in \reffig{inference}, the framework can generate synchronized duets from audio~(\reffig{inference}(a)), predict a partner's motion given one performer~(\reffig{inference}(b)), or refine existing single-singer sequences with interaction-aware adjustments~(\reffig{inference}(c)).

\noindent\textbf{Dyadic Generation.}
Starting from pure Gaussian noise, the model progressively denoises both performers' motion sequences conditioned on separated audio features and interaction scores. Since neither motion stream is predefined, both singers are synthesized jointly, allowing mutual interaction patterns to emerge throughout the denoising process. This mode generates fully synchronized duet performances directly from audio.

\noindent\textbf{Partner Generation.}
When one performer's motion is available, only a small amount of noise is injected into the known motion tokens, while the partner's tokens remain heavily corrupted. During denoising, the model preserves the given performance while generating the missing partner conditioned on both audio and the learned interaction representation, resulting in temporally aligned and complementary motions without requiring retraining.

\noindent\textbf{Score-guided Generation.}
Given a pre-generated single-singer sequence, interaction scores determine the amount of injected noise at each frame. Frames with stronger interaction receive larger perturbations to encourage expressive responses, whereas low-interaction regions remain largely unchanged to preserve the original performance. The diffusion model subsequently refines the sequence into an interaction-aware duet while maintaining the overall motion quality and temporal consistency.

Across all inference modes, the same pretrained model is employed without task-specific fine-tuning or architectural modification. 
By exploiting the flexibility of diffusion forcing, InterSing supports training-free adaptation to different input conditions while maintaining interaction-aware generation.




\section{Experiments}

\begin{table*}[t]
  \centering
    \caption{\textbf{Quantitative comparisons with state-of-the-art methods.}
    \textbf{Bold} indicates the best result, and \underline{underline} indicates the second best.
    $\downarrow$ means lower is better, $\uparrow$ means higher is better, and $\rightarrow$ means closer to the GT is better.
    LVE, FVE are reported in $mm$, and FDD in $1 \times 10^{-2} mm$. 
    }
  \label{tab:comparison}
  \begin{tabular}{@{}c|c|*{8}{c}@{}}
    \bottomrule
    \multicolumn{2}{c|}{Method} & LVE $\downarrow$ & FVE $\downarrow$      & FDD  $\rightarrow$     & Freeze Rate $\downarrow$    & BA $\rightarrow$  &FID $\downarrow$  & P-FID $\downarrow$ \\
    \hline
    \multicolumn{2}{c|}{GT} & -        & -         &0.0000             & -                     &0.2380      & -         & -           \\
    \hline
    \multirow{3}{*}{
    \begin{tabular}{c} Dyadic \\ Generation \end{tabular}}
      & PaChorus \cite{xie2025let}       & 10.9839           & 2.6282               & 1.6824                & 7.2220                     &\underline{0.2487}     & 40.8057            &45.1401   \\
      & UniLS \cite{chu2025unils}        &\underline{9.9350} & \underline{2.3844}   & \underline{0.3624}    & \underline{6.0304}         &0.2251                 & \underline{6.4385} &\underline{9.1933}    \\
      & \cellcolor{gray!20}InterSing (Ours) & \cellcolor{gray!20}\textbf{8.3317} & \cellcolor{gray!20}\textbf{2.0732} & \cellcolor{gray!20}\textbf{0.3212} & \cellcolor{gray!20}\textbf{5.9709} & \cellcolor{gray!20}\textbf{0.2472} & \cellcolor{gray!20}\textbf{4.5566} & \cellcolor{gray!20}\textbf{6.8160} \\
    \cline{1-9}
    \multirow{3}{*}{
    \begin{tabular}{c} Partner \\ Generation\end{tabular}}
      & DIM \cite{tran2024dim}           & 11.2558           &2.6166                &2.1906                 & \underline{10.0697}           &\underline{0.2712}     &34.9875             &18.7729    \\
      & DualTalk \cite{peng2025dualtalk} &\underline{7.9793} &\textbf{1.9055}       &\underline{0.8370}     & 11.5176                       &0.1991                 &\underline{4.7598}  &\underline{3.6627}    \\
      & \cellcolor{gray!20}InterSing (Ours) & \cellcolor{gray!20}\textbf{7.8619} & \cellcolor{gray!20}\underline{1.9792} & \cellcolor{gray!20}\textbf{0.0149} & \cellcolor{gray!20}\textbf{5.7853} & \cellcolor{gray!20}\textbf{0.2491} & \cellcolor{gray!20}\textbf{4.4513} & \cellcolor{gray!20}\textbf{3.4974} \\
    \toprule
  \end{tabular}
\end{table*}
\begin{table*}[t]
  \centering
    \caption{\textbf{Ablation study on different components in our framework.}
    \textbf{Bold} indicates the best result, and \underline{underline} indicates the second best.
    $\downarrow$ means lower is better, $\uparrow$ means higher is better, and $\rightarrow$ means closer to the GT is better.
    LVE, FVE are reported in $mm$, and FDD in $1 \times 10^{-2} mm$.
    }
  \label{tab:ablation}
  \begin{tabular}{@{}c|c|*{7}{c}@{}}
    \bottomrule
    \multicolumn{2}{c|}{Method} & LVE $\downarrow$ & FVE $\downarrow$ & FDD $\rightarrow$ & Freeze Rate $\downarrow$ & BA $\rightarrow$ & FID $\downarrow$ & P-FID $\downarrow$ \\
    \hline
    \multirow{3}{*}{\rotatebox[origin=c]{90}{\scriptsize\bfseries Stage I}}
      & w/o $\mathcal{L}_{inter}$        & 10.4212           & 2.3985            & 0.6701            & 11.7452           & 0.2423            & 10.5597            & 10.8571 \\
      & w/o $\mathcal{L}_{intra}$        & 8.8970            & 2.1960            & 0.3514            & 6.2953            & 0.2461            & 9.1012            & 9.6027 \\
      & w/o neck mask                    & 8.3711            & 2.0950            & 0.3223            & 7.0501            & \textbf{0.2414}   & 7.9962            & 10.1200 \\
    \hline
    \multirow{3}{*}{\rotatebox[origin=c]{90}{\scriptsize\bfseries Stage II}}
      & w/o inter attn                   & 8.6504            & 2.1342            & \textbf{-0.1877}  & \textbf{4.8153}   & \underline{0.2435}& 11.2109           & 13.3974 \\
      & w/o score                        & 9.0536            & 2.1820            & 0.4610            & 9.9407            & 0.2480            & 5.5752            & 7.8203 \\
      & GT score                         & \textbf{8.2576}   & \textbf{2.0575}   & \underline{0.2955}& 5.9978            & 0.2499            & \textbf{4.5220}   & \textbf{6.6525} \\
    \hline
    \multicolumn{2}{c|}{\cellcolor{gray!20}\textbf{Ours}} & \cellcolor{gray!20}\underline{8.3317} & \cellcolor{gray!20}\underline{2.0732} & \cellcolor{gray!20}0.3212 & \cellcolor{gray!20}\underline{5.9709} & \cellcolor{gray!20}0.2472 & \cellcolor{gray!20}\underline{4.5566} & \cellcolor{gray!20}\underline{6.8160} \\
    \toprule
  \end{tabular}
\end{table*}

\subsection{Setting}
\subsubsection{Dataset}

We adapt ChorusHead~\cite{xie2025let}, the only available 3D chorus singing dataset to date, for interaction-aware duet animation.
Since the original dataset focuses on collective chorus representation rather than pairwise interaction modeling, we introduce a targeted preprocessing pipeline to extract reliable dyadic singing sequences and construct interaction-aware supervision.

We first curate the data by removing unsuitable samples, including pseudo-duet cases such as single-person role switching and non-interactive back-to-back performances, as well as unstable side-profile reconstructions that introduce unreliable motion estimation.
We further supplement the dataset with real duet singing videos from publicly available sources to ensure sufficient scale and increase the diversity of interpersonal coordination patterns.
This step is necessary because ChorusHead, although providing a valuable foundation, is not designed for interaction-aware duet generation: over 75\% of its frames correspond to face-to-face interaction according to the relative neck orientation criterion described in Sec.~\ref{sec:stage1}, while common patterns such as one-sided gaze and weak/no-interaction segments are underrepresented, which may bias models toward the dominant interaction mode.

For motion reconstruction, we replace the original center-line frame-splitting strategy with direct MTCNN~\cite{zhang2016joint} detection on full two-person frames, yielding substantially more coherent 3D head motion sequences.
Audio-wise, we employ the AV-MossFormer2-TSE-16K pretrained model~\cite{zhao2025clearervoice} to extract individual singer tracks from mixed audio-visual recordings, decoupling generation quality from separation fidelity.
We further leverage an active speaker detection model to derive per-frame speaker activity scores as an auxiliary condition.

The resulting dataset contains 9 hours of duet-singing performances, including 414 clips, 149 songs, and 972,394 frames at 30 FPS.
It covers 61 singers (41 female and 20 male performers) from 39 groups, spanning diverse musical styles including R\&B, pop, rock, and other genres.
Approximately 90\% of the songs are in English, with the remaining samples covering languages such as Chinese and Japanese.
For evaluation, we adopt a song-disjoint 85:15 train/test split, ensuring that no songs overlap between training and testing sets.

\subsubsection{Implementation Details}
In Stage I, the interaction encoder is optimized by AdamW~\cite{loshchilov2017decoupled} with a learning rate of $2.0 \times 10^{-4}$ and a batch size of 128 on 2 NVIDIA RTX 4090 GPUs.
The interaction modeling process uses a neck-rotation threshold $\phi = 22.5^\circ$, a temporal distance threshold $d = 15$ frames, and a temperature coefficient $\tau = 0.07$.
In Stage II, the generation model is trained with AdamW using a learning rate of $4.0 \times 10^{-4}$ and the same batch size on 2 NVIDIA RTX 4090 GPUs, with loss weights set to $\lambda_1 = 1$, $\lambda_2 = 3$, $\lambda_3 = 2$, $\lambda_4 = 2$, and $\lambda_5 = 0.5$.
During inference, our method runs at over 250 FPS on a single NVIDIA RTX 4090 across all three inference modes.

\subsubsection{Evaluation Metrics}
Following existing works~\cite{fan2022faceformer, xing2023codetalker, stan2023facediffuser, peng2023selftalk, thambiraja2023imitator, huang2025think2sing, chu2025unils}, we evaluate the generated results from five aspects: facial accuracy, head pose accuracy, rhythm alignment, temporal coherence and quality, and interaction consistency. For facial accuracy, Lip Vertex Error (LVE), Facial Dynamic Distance (FDD), Facial Vertex Error (FVE), and Freeze Rate are adopted to evaluate lip synchronization, upper facial dynamics, and overall facial geometry, respectively. 
Rhythm alignment is evaluated by Beat Alignment (BA), which measures the consistency between singing-driven head motion and musical beats.
Temporal coherence and motion quality are quantified using Fréchet Inception Distance (FID) computed on FLAME parameters.
Interaction consistency is measured by Paired Fréchet Inception Distance (P-FID), which compares the distribution of generated and real dyadic motions to assess whether realistic interpersonal coordination is preserved.

\subsection{Comparison with Existing Methods}
The work most directly related to our task is PaChorus~\cite{xie2025let}, which generates head animations of both performers from duet singing audio.
Given the scarcity of duet-singing head animation methods, we further adopt representative conversational interaction approaches as additional baselines, since they also model inter-participant motion dynamics.
Specifically, DualTalk~\cite{peng2025dualtalk} and DIM~\cite{tran2024dim} condition the generation process on the motion sequence of the counterpart in addition to the audio signals, while UniLS~\cite{chu2025unils} jointly synthesizes head motions from the audio streams of both participants.
All baselines are retrained on the same dataset for fair comparison.

\subsubsection{Quantitative Results}
\reftab{tab:comparison} presents quantitative comparisons on both the Dyadic Generation and Partner Generation tasks. 
All baselines are retrained on our dataset using their official or author-provided implementations for fair comparison.
Crucially, a single checkpoint of our method is evaluated across both settings without task-specific fine-tuning, owing to the decoupled interaction conditioning that enables multi-task inference within a unified model.
For dyadic generation, InterSing achieves comprehensive superiority over all baselines across every metric.
Most strikingly, it reduces LVE by \textbf{16.1\%} relative to the strongest baseline UniLS (8.33 vs.\ 9.94), reflecting substantially more precise lip articulation.
The FID/P-FID relative reductions of \textbf{29.2\%}/\textbf{25.9\%} further confirm that the generated distributions closely approximate the ground-truth manifold, validating the effectiveness of our interaction-aware formulation in preserving both individual expressiveness and dyadic coherence.
Notably, PaChorus exhibits a substantially higher FID, likely because it generates overly similar motions for both performers without explicit coordination modeling, leading to a less realistic duet motion distribution.
For partner generation, while DualTalk attains a marginally lower FVE, this advantage comes at the cost of severely degraded temporal dynamics.
Its Freeze Rate and BA reveal a collapse toward temporally averaged, expressively impoverished motions.
In contrast, InterSing attains an order-of-magnitude lower FDD with the lowest Freeze Rate, recovering fine-grained facial dynamics at substantially higher fidelity while maintaining superior motion diversity, as further corroborated by consistent FID, P-FID, and BA gains.

\begin{figure*}[t]
  \centering
  \includegraphics[width=\linewidth]{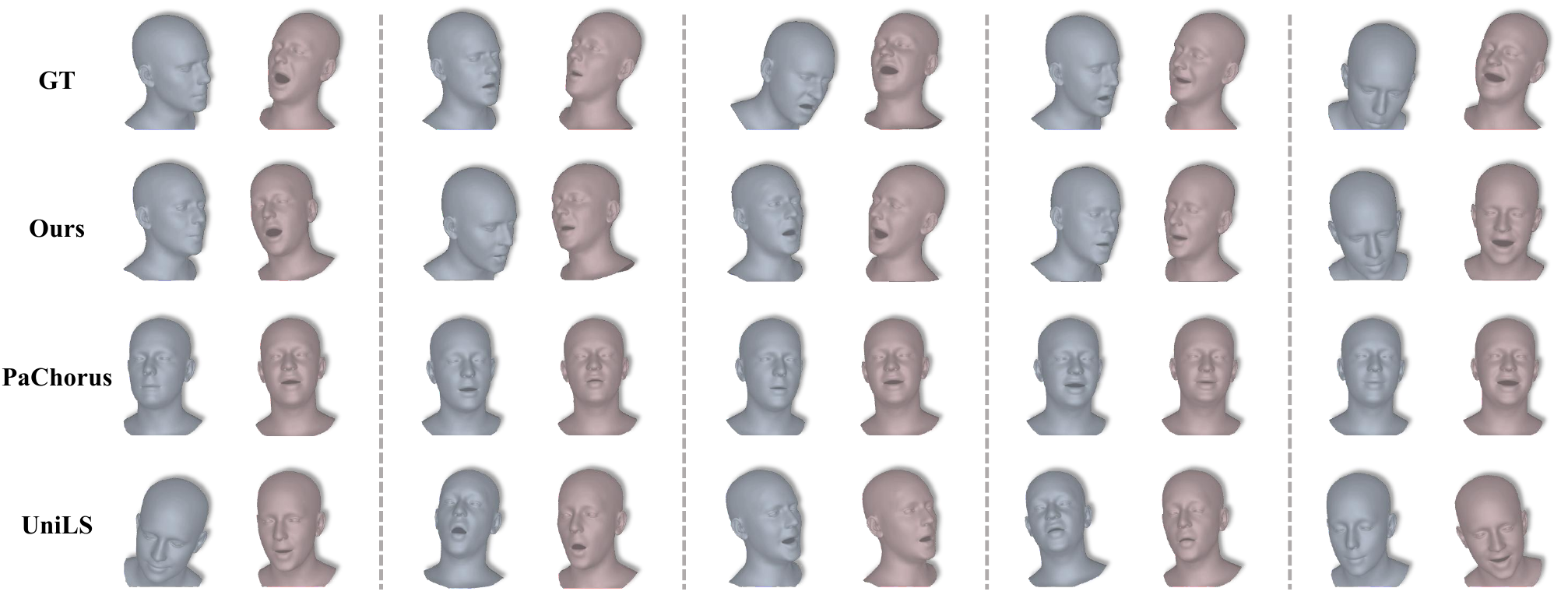}
    \caption{\textbf{Qualitative comparison for dyadic generation.} Compared with existing methods, our approach generates duet animations that better match the ground truth in lip synchronization, facial expression dynamics, and mutual interaction behaviors between performers.}
  
  \label{comparison_dyalic}
\end{figure*}
\begin{figure*}[t]
  \centering
  \includegraphics[width=\linewidth]{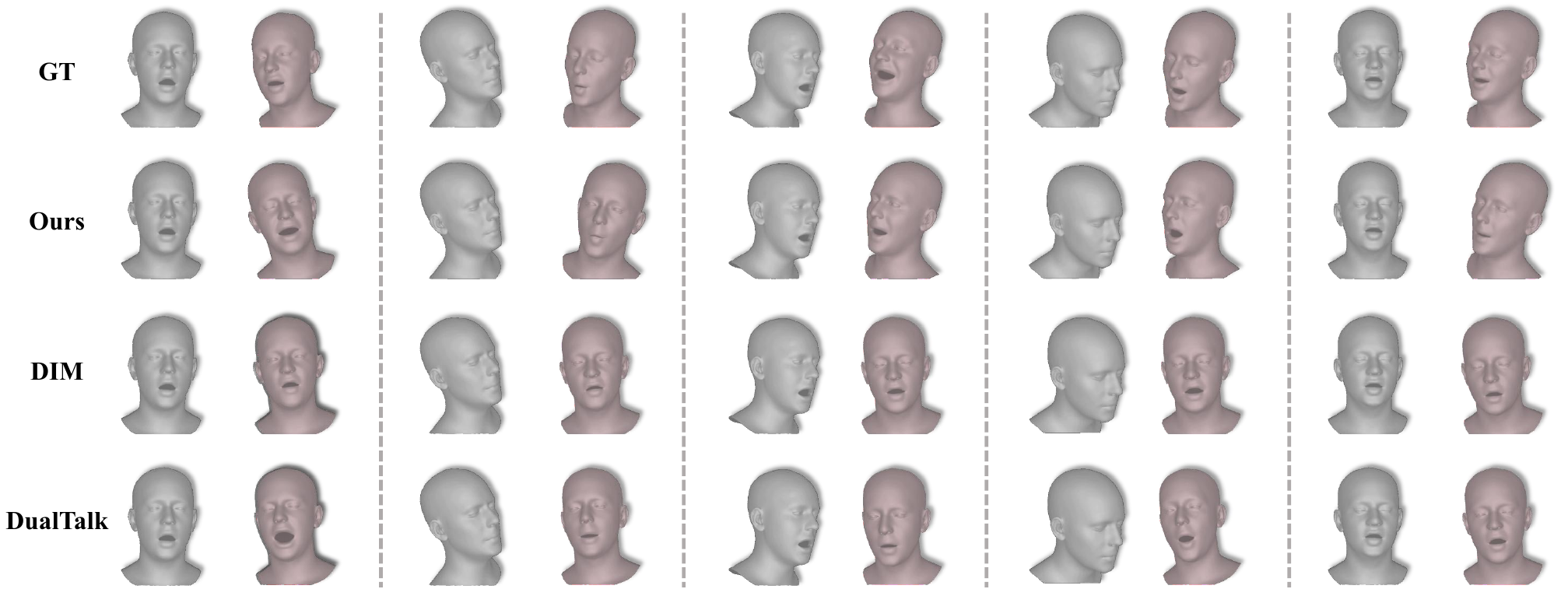}
  \caption{\textbf{Qualitative comparison for partner generation.} Given the reference motion of one performer (shown in gray), our method generates more coherent and interaction-aware responses than competing methods, while better preserving lip synchronization, expression continuity, and responsive head dynamics.}
  
  \label{comparison_partner}
\end{figure*}
\subsubsection{Qualitative Results}
\begin{figure*}[t]
  \centering
  \includegraphics[width=\linewidth]{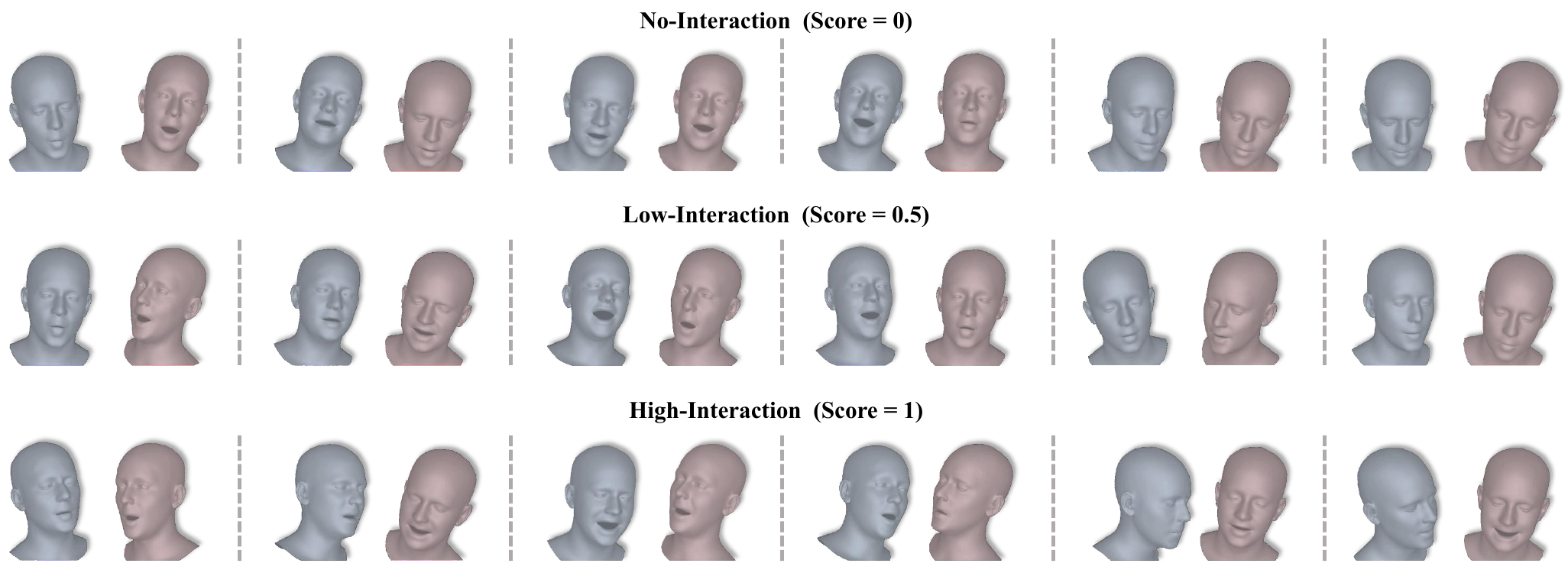}
    \caption{\textbf{Visualization of score control.} Facial motions generated from the same audio input under no-interaction, low-interaction and high-interaction score conditions.}
  
  \label{score_control}
\end{figure*}
\begin{figure*}[t]
  \centering
  \includegraphics[width=0.85\linewidth]{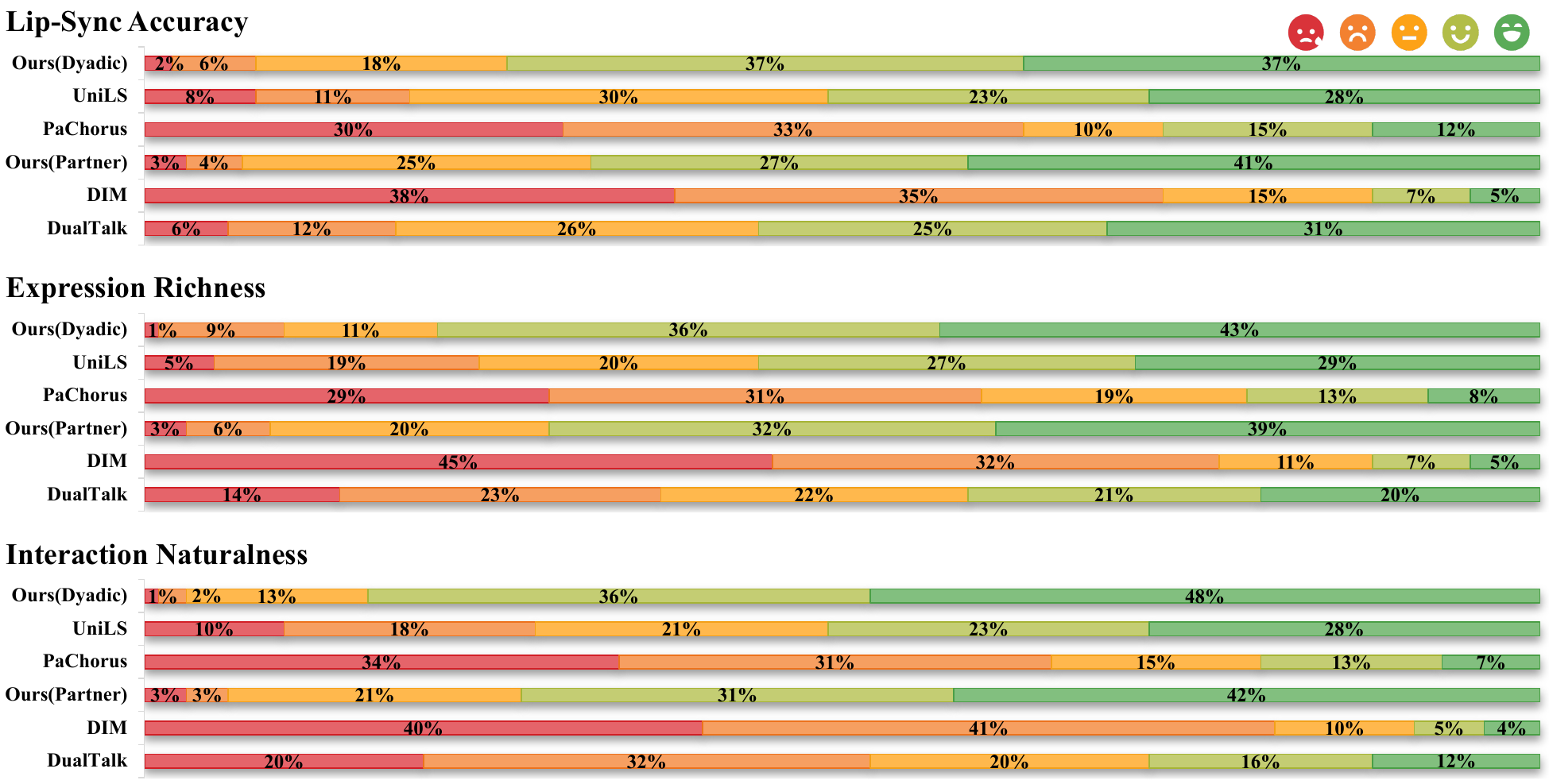}
    \caption{\textbf{User study on lip-sync accuracy, expression richness, and interaction naturalness.} Our method consistently achieves the highest user preference rates across all evaluation criteria.}
  
  \label{userstudy}
\end{figure*}
We present qualitative comparisons in \reffig{comparison_dyalic} and \reffig{comparison_partner}, where existing methods often produce relatively mechanical and neutral facial animations, lacking the expressive variations and harmonious coordination required for duet singing. In contrast, InterSing generates more natural interaction patterns between singers, with better synchronization in facial movements and more coherent responses. \reffig{score_control} further illustrates motion variations under identical audio with different interaction score conditions, demonstrating that the generated animations can adapt to different levels of interaction while maintaining consistent audio-driven dynamics.
To further assess perceptual quality, we conducted a user study with 28 participants (16 males and 12 females, aged between 20 and 45 years old). Participants were presented with 40 randomly ordered animation samples (20 samples were selected from the dyadic generation task and 20 samples from the partner generation task), where the identities of the compared methods were concealed to avoid potential bias. For each sample, animations generated by different methods were rendered under the same audio input and displayed in a randomized order. Participants evaluated the generated singing animations on a 1--5 Likert scale according to three criteria: lip-sync accuracy, expression richness, and interaction naturalness. Lip-sync accuracy measures the synchronization between facial motion and singing audio, expression richness evaluates the diversity and intensity of facial expressions, and interaction naturalness assesses the perceived coordination and responsiveness between performers. As shown in \reffig{userstudy}, InterSing achieves the highest scores across all perceptual dimensions, demonstrating its ability to generate more synchronized, expressive, and socially coherent duet singing animations.

\subsection{Ablation Study}
We conducted ablation experiments to evaluate the contribution of each component, with results summarized in~\reftab{tab:ablation}.
For Stage I, the absence of $\mathcal{L}_{inter}$ causes the most severe degradation across the board, with substantial increases in both distributional metrics and Freeze Rate, confirming that without explicit discrimination between interaction and non-interaction states, the encoder produces an ambiguous latent space incapable of guiding temporally stable generation.
$\mathcal{L}_{intra}$ proves similarly critical, as its removal inflates FID and P-FID, validating that temporal contrastive alignment is essential for resolving fine-grained coherence within interaction segments.
Disabling neck masking preserves per-frame accuracy yet degrades P-FID, indicating that the encoder shortcuts through low-level motion cues rather than distilling genuine interaction semantics.
These findings collectively confirm that the representational quality of Stage I directly governs downstream generation fidelity, with $\mathcal{L}_{intra}$ and $\mathcal{L}_{inter}$ serving as complementary regularizers that enforce temporal coherence and interaction discriminability, respectively.
For Stage II, replacing interaction cross-attention with naive feature concatenation yields the worst P-FID among all variants, confirming that without explicit cross-person dependency modeling the generator produces independently plausible yet interactionally incoherent motions.
Removing score injection preserves competitive FID/P-FID yet causes Freeze Rate and FDD to surge, exposing that cross-attention alone suffices for local realism but lacks the global temporal scaffold that the interaction score provides for long-range coordination.
This complementarity is further corroborated when predicted scores are replaced with ground-truth ones, yielding the best FID and P-FID, which confirms that more precise interaction priors translate directly into more stable diffusion guidance and validates the controllability of score conditioning for multi-person generation.
\section{Extension to Multi-Person Scenarios}
\begin{figure*}[t]
  \centering
  \includegraphics[width=\linewidth]{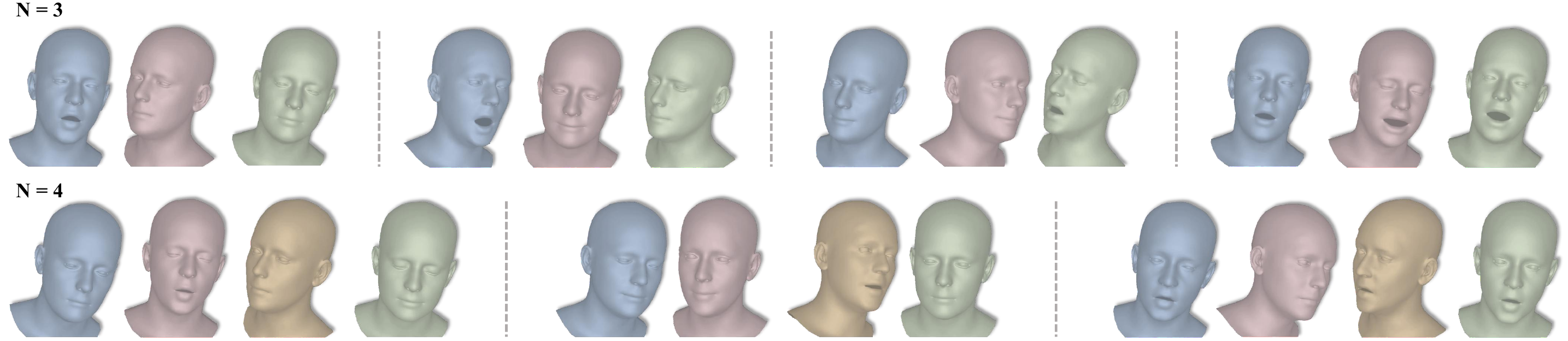}
    \caption{\textbf{Visualization of multi-person chorus generation.} By decomposing global coordination into sequential pairwise interactions, InterSing extends training-free to multi-person singing, preserving coherent interpersonal interactions.}
  
  \label{application}
\end{figure*}
\begin{table}[t]
  
\begin{minipage}{\linewidth}
    \centering
    \caption{\textbf{Quantitative evaluation on the collected three-singer subset.}
        \textbf{Bold} indicates best result. $\downarrow$ means lower is better, and $\rightarrow$ means closer to the GT is better.
    }
        {
            \setlength{\tabcolsep}{3.5pt}
            \renewcommand\arraystretch{1.05}
            \begin{tabular}{c|ccccc} \hline
                Methods                   & LVE $\downarrow$       & FVE $\downarrow$     & FDD $\rightarrow$    & Freeze Rate $\downarrow$  & BA $\rightarrow$\\ \hline
                GT                        & -                      & -                    & 0.0000               & -                         & 0.2304    \\
                PaChuros                  & 11.2342                & 2.7491               & 1.1570               & 7.7144                    & 0.2520  \\
                \cellcolor{gray!20}InterSing (Ours)          &\cellcolor{gray!20}\textbf{8.8769}                 &\cellcolor{gray!20}\textbf{2.2356}               &\cellcolor{gray!20}\textbf{0.3375}          &\cellcolor{gray!20}\textbf{5.5617}                 &\cellcolor{gray!20}\textbf{0.2413} \\ \hline
            \end{tabular}
            \label{tab:application}
    }
\end{minipage} 

\end{table}

The decoupled interaction modeling and score-conditioned generation of InterSing naturally lend themselves to a training-free extension beyond the duet setting, as shown in~\reffig{application}. By arranging avatars horizontally with interactions restricted to immediate neighbors, global multi-person coordination decomposes into sequential pairwise generation steps that directly reuse the pretrained dyadic framework.
For three participants, the motions of A and B are first synthesized jointly via Dyadic Generation. With B's motion fixed, the B--C pair is produced through score-guided Partner Generation under a mutual exclusion constraint, which prevents the B--C interaction score from occupying temporal regions already assigned to the A--B interaction during noise injection, thereby reducing conflicts between different interaction relationships.
This formulation generalizes to arbitrary participant counts through a singing visual center that determines generation order. Central interactions are synthesized first, with generation expanding progressively outward, each peripheral interaction score constrained by already-established central ones. This prioritization grants higher stability to visually dominant participants while ensuring peripheral motions adapt coherently to the global interaction structure.

To validate the composability of the proposed interaction representation in multi-singer scenarios, we collected 30 minutes of three-singer performances and compared InterSing with PaChorus~\cite{xie2025let}. Unlike PaChorus, which represents all non-target singers as a single partner representation and therefore models only coarse one-to-many interactions, our framework explicitly captures pairwise interaction scores that can be composed to represent multiple interaction relationships simultaneously. As summarized in~\reftab{tab:application}, InterSing consistently outperforms PaChorus on all metrics, demonstrating that explicit pairwise interaction modeling provides a more effective foundation for extending duet singing animation to multi-singer scenarios. 
These results further verify that the learned interaction representation is compositional and can be flexibly reused to model complex multi-person coordination without additional training.
\section{limitation}
Similar to other FLAME-based methods~\cite{xie2025let, peng2025dualtalk}, our framework does not explicitly capture fine-grained eye-gaze dynamics, due to the inherent limitations of the FLAME representation. Another limitation is the lack of large-scale multi-person chorus datasets, which prevents comprehensive quantitative evaluation in multi-singer scenarios. In the future, we will explore more expressive 3D facial representations with richer eye and gaze modeling capabilities, and investigate large-scale multi-person singing datasets to enable more thorough evaluation and further improve the scalability of interactive singing animation.

\section{Conclusion}

We present InterSing, a framework for 3D multi-person singing animation that explicitly models the intermittent, rhythm-driven interaction dynamics in duet performances. Moving beyond conventional joint-generation paradigms and implicit interaction modeling, we reformulate duet head animation as an interaction-conditioned motion generation problem.
Our approach consists of a two-stage pipeline. In Stage I, a weakly supervised contrastive learning scheme, guided by geometric pseudo-labels and enhanced with a neck-masking strategy, learns a time-varying and interpretable interaction signal that captures coordination strength beyond raw head orientation cues. In Stage II, an interaction-aware diffusion transformer conditioned on audio, speaker activity, and the learned interaction signal generates temporally coherent and mutually responsive dyadic motions.
By incorporating diffusion forcing, a single unified model supports multiple downstream tasks, including full duet synthesis, partner motion prediction, and interaction-guided refinement. Extensive quantitative evaluations and a user study show that InterSing consistently outperforms state-of-the-art baselines in lip synchronization, facial expressiveness, and perceived interaction realism. Ablation studies further confirm the importance of explicit interaction modeling and the proposed contrastive learning design.

\ifCLASSOPTIONcaptionsoff
  \newpage
\fi

\bibliographystyle{IEEEtran}
\bibliography{main}
 
%


\begin{IEEEbiography}[{\includegraphics[width=1in,height=1.25in,clip,keepaspectratio]{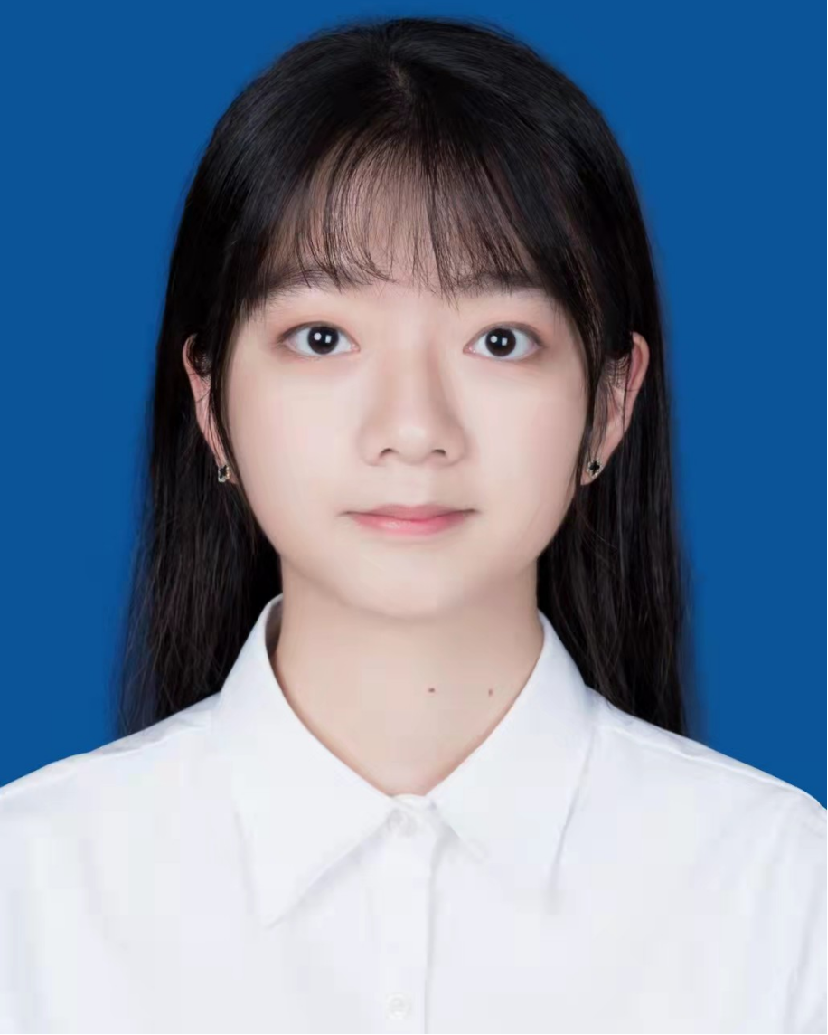}}]{Yihan Zhou} is a M.S. student in the School of Computer Science and Engineering, South China University of Technology. Her research interests include computer vision, computer graphics and multimodal learning.
\end{IEEEbiography}

\begin{IEEEbiography}[{\includegraphics[width=1in,height=1.25in,clip,keepaspectratio]{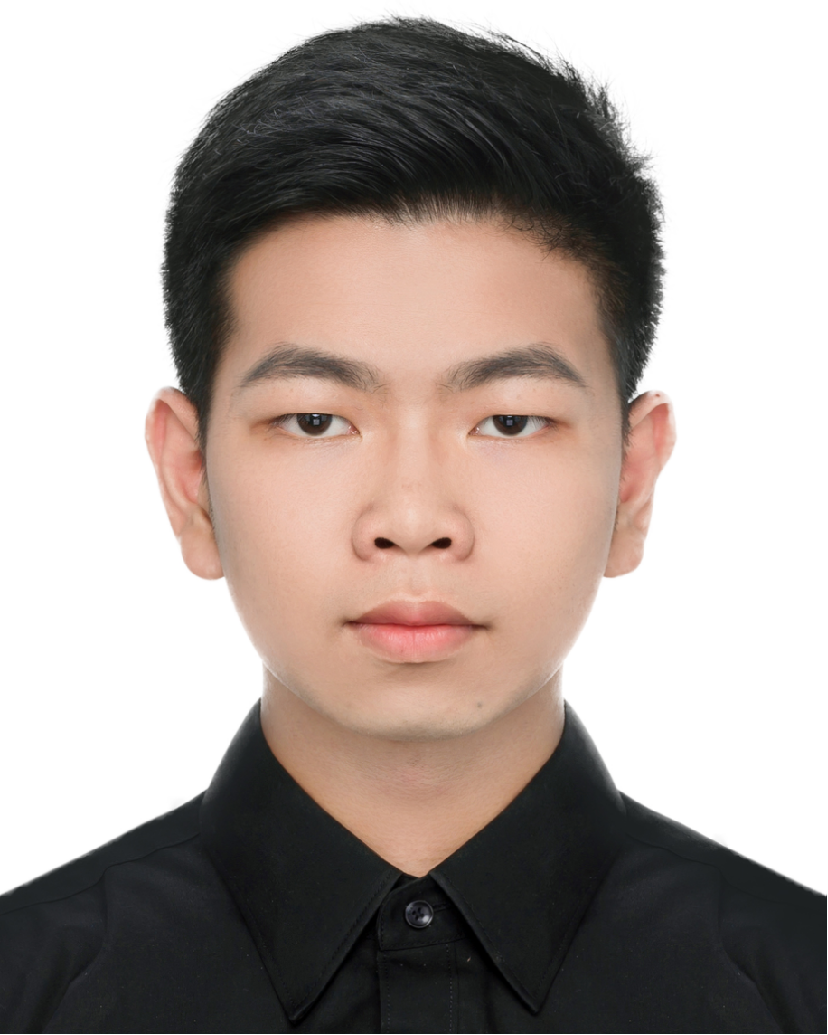}}]{Zikai Huang} is currently pursuing a Ph.D. degree in the School of Computer Science and Engineering, South China University of Technology. His research interests include computer vision, computer graphics and multimodal learning.
\end{IEEEbiography}

\begin{IEEEbiography}[{\includegraphics[width=1in,height=1.25in,clip,keepaspectratio]{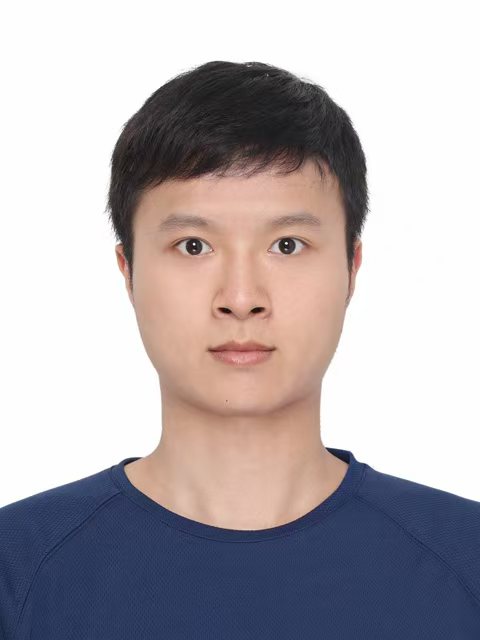}}]{Yuyang Yu} received his B.Sc. degree in Communication Engineering from Jinan University in 2021. He is currently working toward the Ph.D. degree in the School of Computer Science and Engineering, South China University of Technology. His research interests include Multimodal Learning, Visual Anomaly Detection.
\end{IEEEbiography}

\begin{IEEEbiography}[{\includegraphics[width=1in,height=1.25in,clip,keepaspectratio]{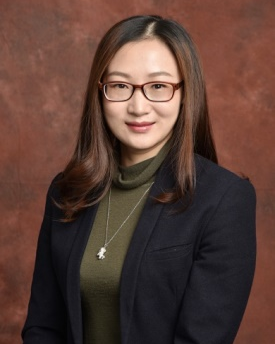}}]{Xuemiao Xu} received the BS and MS degrees in computer science and engineering from South China University of Technology, in 2002 and 2005, respectively, and the PhD degree in computer science and engineering from The Chinese University of Hong Kong in 2009. She is currently a professor with the School of Computer Science and Engineering, South China University of Technology. Her research interests include object detection, tracking, recognition, and image, video understanding and synthesis, particularly their applications in the intelligent transportation.
\end{IEEEbiography}

\begin{IEEEbiography}[{\includegraphics[width=1in,height=1.25in,clip,keepaspectratio]{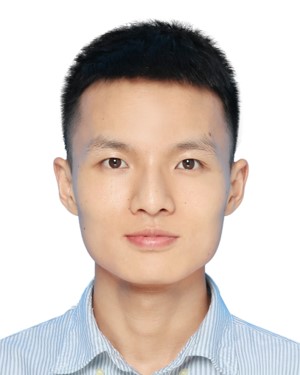}}]{Cheng Xu} received his Ph.D. in Computer Science and Technology from South China University of Technology, China, in 2023. He is currently a Research Scientist at Singapore Management University. Prior to this, he was a Post-Doctoral Fellow at The Hong Kong Polytechnic University from 2023 to 2026. His research interests primarily include human-centric visual representation, understanding, and generation.
\end{IEEEbiography}

\begin{IEEEbiography}[{\includegraphics[width=1in,height=1.25in,clip,keepaspectratio]{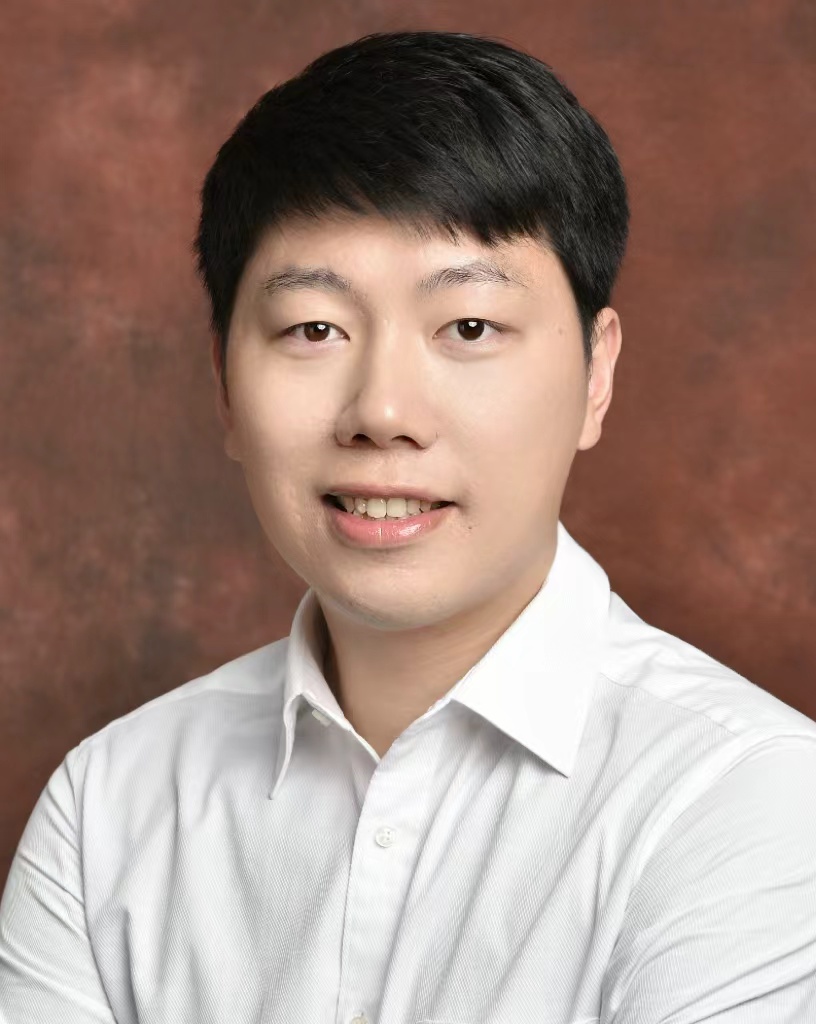}}]{Shengfeng He} (Senior Member, IEEE) is an associate professor in the School of Computing and Information Systems at Singapore Management University. Previously, he was a faculty member at South China University of Technology. He earned his B.Sc. and M.Sc. from Macau University of Science and Technology and a Ph.D. from City University of Hong Kong. His research focuses on computer vision and generative models. He has received awards including the Google Research Award, PerCom 2024 Best Paper Award, and the Lee Kong Chian Fellowship. He is a senior IEEE member and distinguished CCF member. He serves as lead guest editor for IJCV and associate editor for IEEE TPAMI, IEEE TNNLS, IEEE TCSVT, Visual Intelligence, and Neurocomputing. He is an area chair/senior PC member for CVPR, NeurIPS, ICLR, ICML, AAAI, IJCAI, and BMVC, and will serve as Conference Chair of Pacific Graphics 2026.
\end{IEEEbiography} 
 




\vfill

\end{document}